\documentclass{article}
\usepackage[]{neurips_2019}
\usepackage[ruled,vlined]{algorithm2e}
\usepackage{ amssymb }
\usepackage{amsmath}
\usepackage{ dsfont }
\usepackage{xcolor}
\usepackage{rotating}
\usepackage{booktabs}
\usepackage{graphicx}
\usepackage[export]{adjustbox}
\usepackage{url}
\usepackage{hyperref}

\DeclareMathOperator*{\argmax}{arg\,max}

\DeclareMathOperator*{\ComputeFitness}{ComputeFitness}
\DeclareMathOperator*{\Softmax}{Softmax}
\DeclareMathOperator*{\odds}{odds}
\DeclareMathOperator*{\score}{score}
\title{PermuteAttack: Counterfactual Explanation of Machine Learning Credit Scorecards}

\author{
	Masoud Hashemi \thanks{All contents and opinions expressed in this document are solely those of the authors and do not represent the
view of RBC Financial Group.} \\
	Borealis AI\\
	777 Bay St. Toronto \\
	\texttt{masoud.hashemi@borealisai.com} \\
	\And
	Ali Fathi \footnotemark[1]\\
	Royal Bank of Canada (RBC) \\
	200 Bay St. Toronto \\
	\texttt{ali.fathi@rbc.com} \\
}

\begin{document}

\maketitle

\begin{abstract}
This paper is a note on new directions and methodologies for validation and explanation of Machine Learning (ML) models employed for retail credit scoring in finance. Our proposed framework draws motivation from the field of Artificial Intelligence (AI) security and adversarial ML where the need for certifying the performance of the ML algorithms in the face of their overwhelming complexity poses a need for rethinking the traditional notions of model architecture selection, sensitivity analysis and stress testing. Our point of view is that the phenomenon of adversarial perturbations, when detached from the AI security domain, has purely algorithmic roots and fall within the scope of model risk assessment. We propose a model criticism and explanation framework  based on \textit{adversarially generated counterfactual examples } for tabular data. A counterfactual example to a given instance in this context, is defined as a synthetically generated data point sampled from the estimated data distribution which is treated \textit{differently} by a model.
 
The counterfactual examples can be used to provide black-box instance level explanation of the model behaviour as well as studying the regions in the  input space where the model performance deteriorates.  Adversarial example generating algorithms are extensively studied in image and natural language processing  (NLP) domains. However, most financial data come in tabular format and naive application of the existing techniques on this class of datasets generates unrealistic samples. In this paper we propose a counterfactual example generation method capable of handling tabular data including discrete and categorical variables. Our proposed algorithm uses a gradient free optimization based on genetic algorithms and therefore is applicable to any classification model.
\end{abstract}

\section{Introduction}
The recent revolutionary advances in deep learning applied to computer vision, speech recognition, natural language processing and reinforcement learning has ignited motivation in quantitative finance research for applying similar techniques in areas such as asset pricing, trading credit risk, algorithmic trading, fraud modelling and retail credit risk scoring \cite{Collaris2018InstanceLevelEF, Gu2018EmpiricalAP, HenryLabordre2017DeepPA, Ning2018DoubleDQ, Sirignano2016DeepLF}.
Witnessing the success of these algorithms in capturing the relevant latent feature representations of the data relevant to the task at hand, without the need for feature engineering by human, it is natural to expect the same to be accomplished for the typical tabular datasets arising in the financial modelling domain, i.e., to replace the traditional subject matter expert-led model development process by a purely data-driven machine learning process. 

The financial industry has already witnessed significant integration of AI within the advanced analytics functions such as customer engagement via NLP-powered chat bots, marketing, fraud risk management and many other areas. However, at the same time the arising risk as a result of deploying this new generation of analytics products and technologies has already peeked the regulatory interests. For instance, in \cite{osfi2}, OSFI --the Canadian financial industry regulator-- notes that: \textit{"...At the same time, AI presents challenges of transparency and explainability, auditability, bias, data quality, representativeness and ongoing data governance. There are challenges in terms of overall model risk management controls such as continuously evolving models and the use of AI in validation. There may also be risks that are not fully understood and limited time would be available to respond if those risks materialize."} 

The above mentioned considerations are of course not new among the statistics and machine learning communities. For instance, Brieman, in his famous ``Two Cultures'' paper \cite{Breiman2001StatisticalMT} which was published in the early days of ``data science'', already points to the fundamental tensions between the \textit{data modelling culture} of statisticians in contrast with the \textit{algorithmic modelling culture} in the machine learning community. He argues therein that statistical methodology is primarily focused on choosing of an understandable model and studying the resulting derived properties than assuring that the chosen model is a good approximation of the underlying date generating process.  The paper also argues that the statistician's approach to  model validation is done as a yes-no matter using \textit{goodness-of-fit} tests and \textit{residual analysis} and based on this approach, the models cannot be compared regarding with how well they approximate the problem under study. 

To make this point more concrete, one can also observe this limitation of statistical model diagnostics checking through the lens of \textit{Empirical Indistinguishably} and \textit{Fragility} \cite{Wasserman00therole}. An assumption about a probability distribution is called fragile if the assumption holds for a distribution $\mathcal{P}$ but the assumption fails for a distribution $\mathcal{Q}$ where $\mathcal{Q}$ is empirically indistinguishable from $\mathcal{P}$. It is said that $\mathcal{Q}$ is empirically indistinguishable from $\mathcal{P}$ if there is no reliable hypothesis test for: $H_0: X_1,\cdots, X_n\sim\mathcal{P}$ versus  $H_1: X_1,\cdots, X_n\sim\mathcal{Q}$.

Taking the Total Variation distance ($TV$) as the metric for proximity of two distributions,
\begin{equation}
TV(\mathcal{P},\mathcal{Q}) = \sup_{A}|\mathcal{P}(A)-\mathcal{Q}(A)|
\end{equation}
it can be shown that for each distribution $\mathcal{P}$, the neighbourhood,
\begin{equation}
B_n(\mathcal{P})=\{\mathcal{Q}, TV(\mathcal{P},\mathcal{Q})\leq\frac{\epsilon^2}{4n}\}
\end{equation}
is distinguishable from $\mathcal{P}$ (see \cite{Wasserman00therole} and references therein). It is argued in \cite{Wasserman00therole} that properties such as linearity, parameter sparsity, error constant variance are fragile (more pronounced in the high dimensional problems) and hence, statistical diagnostic tests for checking them may not be reliable.

In the algorithmic modelling (ML practitioners) camp, the widely accepted principle is that accuracy generally requires more complex prediction methods and simple and interpretable functions do not make the most accurate predictors.
While the lack of transparency as a result of increased complexity of the machine learning models is often pointed out, the provided solutions for interpretation of the model outputs (- as a means for establishing trust) seem to fail to give a coherent solution, as pointed out in  \cite{Lipton2018TheMO}:\textit{"...And yet the task of interpretation appears underspecified. Papers provide diverse and sometimes non-overlapping motivations for interpretability, and offer myriad notions of what attributes render models interpretable. Despite this ambiguity, many papers proclaim interpretability axiomatically, absent further explanation."} (also see \cite{Collaris2018InstanceLevelEF} for a case study).

Our conceptual framework is primarily influenced by the point of view proposed in \cite{Wachter2017CounterfactualEW} which is approaching the problem of ML interpretability in the context of the \textit{“right to explanation”} in the EU General Data Protection Regulation (GDPR). In Recital 71 of the GDPR \cite{gdpr}, an indication of the regulatory requirements for implementation of automated decisioning systems is given as follows: \textit{“...should include specific information to the data subject and the right to obtain human intervention, to express his or her point of view, to obtain an explanation of the decision reached after such assessment and to challenge the decision.”}

In \cite{Wachter2017CounterfactualEW}, it is argued that given the significance of protections and rights for individuals in GDPR, the purposes for explanations of an algorithm outcome must be determined from the perspective of the data subject. Hence, they propose three objectives for explanations of automated decisions:\textit{
\begin{itemize}
\item "to inform and help the subject understand why a particular decision was reached, 
\item to provide grounds to contest adverse decisions, and 
\item to understand what could be changed to receive a desired result in the future, based on the current decision-making model."
\end{itemize}}

Therefore, in \cite{Wachter2017CounterfactualEW}, they counterfactual examples are proposed as a means to provide explanations for algorithmic decisions satisfying the above requirements.They argue that while common ML explanation algorithms attempt to convey the internal state or logic of an algorithm that leads to a decision, counterfactuals describe the externalities that led to that decision. Counterfactuals are specifically useful when considering the modern ML models with extremely complex architecture, for which conveying this state to a non-specialist in a way that allows them to reason about the behaviour of an algorithm is extremely difficult.

For the case of explaining the results of a credit scoring model to an individual, the counterfactual example may take the form:\textit{"Your credit score is 700. If your income had been \$85,000, with 5 years of credit history, your score would have been 780.”}

Counterfactual examples are human-friendly explanations, because they are contrastive to the current instance and they are selective, meaning they usually focus on a small number of feature changes. 
There are usually multiple different counterfactual explanations. Each counterfactual could tell a different ``story'' of how a certain outcome was reached. One counterfactual might say to change feature \textit{A}, the other counterfactual might say to leave \textit{A} the same but change feature \textit{B}, which is a contradiction. 
However, it can be advantageous to have multiple counterfactual explanations, because then humans can select the ones that correspond to their previous knowledge.

A counterfactual should be as similar as possible to the instance regarding feature values. This requires a distance measure between two instances. The counterfactual should not only be close to the original instance, but should also change as few features as possible. This can be achieved by selecting an appropriate distance measure like the Manhattan distance. The last requirement is that a counterfactual instance should have feature values that are likely. The counterfactual method creates a new instance, but we can also summarize a counterfactual by reporting which feature values have changed. This gives us two options for reporting our results, one can either report the counterfactual instance or highlight which features have been changed between the instance of interest and the counterfactual instance.

\subsection{Contributions}
In this paper, we employ a black box adversarial generation algorithm called PermuteAttack, in order to generate adversarial example for a machine learning classifier trained on tabular data. Our proposed method is based on genetic algorithm optimization and hence does not depend on differentiability assumptions for the underlying model. We propose a post processing algorithm in order to take into account higher order conditional interactions between the features in order to make sure that the counterfactual example is realistic and is in-distribution. We also provide a graphical explanation of the interactions between the features upon perturbations resulting in the instance level counterfactual example. 

\section{Formal Framework for Adversarial Perturbations}
In this section we outline the formal framework for adversarial perturbations.We follow the exposition in \cite{Werpachowski2019DetectingOV}.

Consider an input space $\mathcal{X}\subset\mathbb{R}^d$ and $\mathcal{Y} = \{0,1,\cdot,K-1\}$ the set of labels. 
Assume that there is a distribution $\mathcal{P}$ over $\mathcal{X}$ from which the inputs are sampled, and the class label is determined
by the ground truth function $f^{*}:\mathcal{X}\longrightarrow \mathcal{Y}$. For an input $X$ drawn from $\mathcal{X}$, its
corresponding ground truth class label is given by $Y = f^*(X)$. Consider a deterministic classifier (for instance a functional form learned as a result of a training procedure) classifiers $f:\mathcal{X}\longrightarrow\mathcal{Y}$ . The performance of $f$ can be measured by the zero-one loss: $L(f,x) = I(f(x)\neq f^*(x))$, where $I(x)$ is the indicator function on $\mathcal{X}$.

For a classification task with the data distribution $\mathcal{P}$ and ground truth $f^*$, an adversarial perturbation generator (APG) for a classifier $f$ is a measurable mapping $g: \mathcal{X}\longrightarrow\mathcal{X}$ satisfying the following property:
\begin{itemize}
    \item $g$ preserves the class labels of the samples, that is, $f^*(x) = f^*(g(x))$ almost every where,
\end{itemize}
The above assumption, called the proximity assumption (see \cite{Wesel2017ChallengesIT}), is important in all the implementations of adversarial perturbation. Concretely, in practice, it is reflected in the assumption that the adversarial perturbations staying in the $\epsilon$-neighbourhood of the original data point $x$ (with respect to some metric) and assuming that the resulting label of $g(x)$ is the same as that of $x$ .

The above characterization considers the case of a hard classifier, however, the implementation of adversarial perturbations generalize to soft classifiers (where $\hat{f}(x)\in(0,1)$) and hence to credit scorecards.
Recall that credit scorecards are essentially monotone decreasing transformations of the output scores of the ML classifier considered as Probability of Default ($PD$) mapping the interval $(0,1)$ onto a pre-specified range of credit scores (see \cite{Siddiqi2005CreditRS}). More precisely, Typically in developing a scorecard, one chooses a baseline score e.g. 600. The baseline score is assumed to denote a good-bad odds of default (say 20:1), meaning
\begin{equation}
\odds(x) =\frac{1-\hat{f}(x)}{\hat{f}(x)}= \frac{1-PD}{PD} = 20.
\end{equation}
It is common to assume that certain amount of increase in the score doubles the odds, meaning that for instance, a score of 615 corresponds to the good bad odds of 40:1 and 630 corresponds 80:1 etc. In this case, it is said that the Point to Double Odds (PDO) is 15. On can use a transformation such as the one given below to map the soft classifier outputs (predicted PDs) to the range of scores:
\begin{equation}\label{eqn:creditscore}
\score = 600+\frac{PDO}{\log(2)}*\log(\frac{\odds}{20}).
\end{equation}

Creating adversarial examples for machine learning models is a very well studied area. However, these algorithms are not very well-suited for tabular data, since the gradient-guided small perturbations used in these algorithms could change the distribution of the tabular data. The distribution changes are more pronounced in the presence of discrete and categorical variables. Here, we review some of the well-known adversarial attacks. 

\textbf{Examples of gradient based algorithms}: Prototype guided algorithms \cite{dhur2018explanations, looveren2019interpretable} use an autoencoder, which is trained to reconstruct instances of the training set, to make the perturbation meaningful. In addition, a prototype loss is added, which tries to minimize the $\ell_2$ distance between the counterfactual and the nearest prototype. As a result, the perturbations are guided to the closest prototype, speeding up the counterfactual search and making the perturbations more meaningful as they move towards a typical in-distribution instance. However, similar to all other gradient based algorithms they cannot handle the categorical variables easily. In addition autoencoders are not very effective for the tabular data (more complicated algorithms must be used such as \cite{xu2018synthesizing, xu2019modeling}).

\textbf{Examples of gradient free algorithms}: The zeroth order optimization (ZOO) \cite{Chen_2017} attack builds on the C\&W attack to perform black-box attacks \cite{carlini2016evaluating}, by modifying the loss function such that it only depends on the output of the DNN, and performing optimization with gradient estimates obtained via finite differences. GenAttack \cite{alzantot2018genattack} relies on genetic algorithms, which are population based gradient-free optimization strategies. These algorithms are also based on adding noise as the perturbation which causes unlikely feature values and cannot handle the categorical variables. 

Our proposed counterfactual generation algorithm, called \textit{PermuteAttack}, is inspired by the GenAttack algorithm \cite{alzantot2018genattack}, with a major difference that PermuteAttack relies on permutation as the data perturbation rather than adding small random values. 

\section{Proposed Algorithm}

PermuteAttack is based on gradient-free optimization genetic algorithm. Genetic algorithms are inspired by the natural selection and evolution. The optimization starts from a pool of samples called population, $\mathcal{P}$, which evolve iteratively to increase a fitness function (i.e., the optimization cost function). In each iteration the population (the population in each iteration is called a generation) goes through selection, crossover, and mutation. In the selection step new population are randomly selected based on their fitness (calculated using the fitness function), meaning the samples with lower costs in a minimization objective have higher chance of being selected. The next generation is generated through a combination of crossover and mutation. Crossover combines the features of the two parents in a random order. Mutation applies some perturbations to some of the features which are chosen randomly. \\
The main difference between our algorithm and other counterfactual algorithms is in the way that the samples are perturbed. In the available genetic algorithm based adversarial attacks, the samples are perturbed by a small noise within a $\ell_p$-ball, which creates some unlikely values specially in the case of categorical and ordinal features. PermuteAttack on the other hand perturbs the sample by permuting randomly selected features. The permutation values are selected randomly from the possible values of the feature in the training data. Therefore, the chosen values are always valid and likely, and the probability distribution of the features remains the same in the new generations.

\subsection{Perturbation Model}
Let $\Pi = \left\lbrace \pi_1, \pi_2, \dotsb, \pi_{n!} \right\rbrace $ be a set of n-length vectors each containing a different permutation of the set $\left\lbrace 1, 2, \dotsb, n \right\rbrace $ and $X = \left[ X_1, X_2, \dotsb, X_m \right] \subset \mathbb{R}^{d \times m}$ be the input data with length of $d$ (each row being a sample of the data) and $m$ columns (columns represent the features). We denote $X_{i} \left[ \pi_{l} \right], \pi_l \in \Pi$ as a permutation of $X_i$ meaning its value is randomly selected from the possible values in the column $i$. $f$ represents the model fitted on $(X, Y)$ pairs, where $Y \in \mathcal{Y}, \mathcal{Y} \subset \mathbb{R}^K$ is the target label with $K$ possible classes and $f(x) \subset \mathbb{R}^K$ is the model prediction for input $x = \left\lbrace x_1, \dotsb, x_m \right\rbrace \subset \mathbb{R}^m$. In addition, $x_{perm}^{(i)} =  \left\lbrace x_1, \dotsb, X_i[\pi_l,0], \dotsb, x_m \right\rbrace$ denotes a sample with $i$th column being permuted by $\pi_l$ and the first value being chosen (i.e., first row of the permuted column). 

The goal of PermuteAttack is to find the $x_{perm}$ which has the least number of permuted columns (features), $\delta_{0,max}$, and the amount of change to be minimum within an $\ell_2$-ball of $\delta_{2,max}$. The permuted sample that satisfies these conditions is the counterfactual $x_{cnt}$.

\begin{align} \label{eqn:model}
& \underset{c \in C}{\argmax} f(x_{cnt}) = t \text{ such that } \nonumber \\
& \|x_{orig} - x_{cnt}\|_0 \le \delta_{0,max} \text{ and } \|x_{orig} - x_{cnt}\|_2 \le \delta_{2,max} 
\end{align}

\subsection{PermuteAttack}
The proposed algorithm to solve equation (\ref{eqn:model}) is presented in Algorithm \ref{alg:permuteattack}, and the main functions of the algorithm are described in this section.

\subsubsection{Fitness Function}
\textit{ComputeFitness} evaluates the quality, of each population member. The fitness function should reflect the threat model. PermuteAttack uses the following fitness function:
\begin{align}
\ComputeFitness(\mathbf{x}) = &\|f(x)_t - f(x_{orig})_t\|_2 - \nonumber \\
\rho_0 &\|x_{orig} - x\|_0 - \nonumber \\
\rho_1 &\|x_{orig} - x\|_2
\end{align}
where $f(x)_t$ is the outcome for the target class $t$. $x$'s with higher fitness values are the samples selected in the selection step. 

\begin{algorithm} \label{alg:permuteattack}
	\SetAlgoLined
	\KwResult{Write here the result }
	\KwIn{original example $x_{orig}$, target label $t$, mutation-range $\alpha$, mutation probability $\rho$, population size $d$, $\tau$ sampling temperature, training data $X_{train}$.}
	$\blacktriangleright$ Create initial generation
	\KwOut{Counterfactual example $x_{cnt}$}
	\For{$i \gets 1$ \textbf{to} $d$ in population}{
		$\mathcal{P}_i^0 \gets Permute(x_{orig}, \Pi, \mathds{1})$
	}
	\For{$g \gets 1$ \textbf{to} $G$ Generations}{
		\For{$i \gets 1$ \textbf{to} $d$ in population}{
			$\mathcal{F}_i^{g-1}$ = ComputeFitness($\mathcal{P}_i^{g-1}$)
		}
		$\blacktriangleright$ Find the elite member.
		
		$x_{c} = \mathcal{P}^{g-1}_{\argmax_j \mathcal{F}_j^{g-1}}$
		
		\If{$\argmax_c f(x_{cnt}) == t $} {
			\Return{$x_{cnt}$} $\triangleright$ Found successful example
		}		
		$\mathcal{P}^{g}_{1} = \{x_{cnt}\}$
		
		$\blacktriangleright$ Compute Selection probabilities
		
		$probs \gets$ \textit{softmax}$(\frac{\mathcal{F}^{g-1}}{\tau})$
		
		\For{$i \gets 2$ \textbf{to} $d$ in population}{
			Sample $parent_1$ from $\mathcal{F}^{g-1}$ according to $probs$
			
			Sample $parent_2$ from $\mathcal{F}^{g-1}$ according to $probs$
			
			$child = CrossOver(parent_1, parent_2)$
			
			$\blacktriangleright$ Apply mutations
			
			$child_{mut} = Mutate(child, \Pi, \mathcal{F}_i^{g-1})$
			
			$\blacktriangleright$ Add mutated child to next generation
			
			$\mathcal{P}^{g}_{i} = \{child_{mut}\}$
		}
		$\blacktriangleright$ adaptively update $\alpha$, $\rho$ parameters
		
		$\rho, \alpha = UpdateParameters(\rho, \alpha)$
	}
	\caption{Proposed PermuteAttack Algorithm}
\end{algorithm}

\subsubsection{Selection}
Members with higher \textit{fitness} are more likely to be a part of the next generation while members with lower fitness are more likely to be replaced. The probability of selection for each population member is computed by the \textit{Softmax} of the fitness values to turn them into a probability distribution. In addition, the member with highest fitness among the current generation is guaranteed to become a
member of the next generation, called an elite member.

\subsubsection{Crossover}
The features of the selected parents are randomly swapped.

\subsubsection{Mutation}
Some of the features of the generation members are selected according to their importance in changing the outcome and their value is randomly replaced by one of the possible values of the feature in the training data. The selection probability is calculated by $$\Softmax(\left\lbrace   e_1, e_2, \dotsb, e_m \right\rbrace),$$
where $e_i$ is the changes in the $f(\mathbf{x})_t$ when $i$th feature is changed in previous generation. Therefore, the features with lower effect on the outcome changes are less likely to be selected in the mutation.

\subsubsection{UpdateParameters}
If the conditions applied on $\ell_0$ and $\ell_2$ (represented by $\rho = \{ \rho_0, \rho_1 \}$) are tight and the algorithm is unable to find any counterfactual example, the value of $\rho$ will be decayed by a predefined factor. This relaxes the conditions and enables the algorithm to find examples with more number of permuted features, with higher deviations from the original values. 

\subsubsection{One-hot encoding}
One-hot encoding is one of the most common categorical variable encoding techniques. It adds dummy variables to the data, each of which representing one of the possible values of the corresponding categorical feature. For each sample, all the dummy variables are zero other than the one that represents the value corresponding to the value of the feature in that specific sample. 

Since the permutation of one-hot encoding is meaningless, before applying the permutations all the one-hot encoded features are converted to ordinal. After mutation the new generated samples are converted back to one-hot encoding to be able to compute the fitness.

\subsubsection{Higher order statistics}
The independent feature perturbation ignores the higher order statistical relations among the features. This is a very common assumption in the adversarial perturbation algorithms, which makes the generated adversarial samples unrealistic. Therefore, the adversarial examples can be detected using an out-of-distribution detection algorithm \cite{carlini2017adversarial}. \\

To improve the quality of the adversarial examples, the sampling should be done from a realistic joint data distribution. In the naive PermuteAttack the features are assumed to be independent and each feature is perturbed independently. To address this issue we limit the permutation to more similar values of the feature. 

To achieve this goal all continuous features are discretized into smaller bins (e.g., 5 bins) such that the same number of samples are in each feature bin. Using the discretized data, to permute the $i^{th}$ feature, the potential permutation values are chosen from the samples that all their discretized features other than $i$ feature are the same, $x_j = x'_j, \forall \{j: j \in \{1, ..., m \} \wedge j \neq i \}$. Therefore, sampling is from the conditional probability distribution $p(x_i|x_{-i})$ where $x_{-i}$ means all features other than the $i^{th}$ one. 

The conditional distribution improves the sampling for one feature given all other features. However, in many mutation and crossover iterations more than one feature is permuted. To sample multiple features we can follow the same procedure. For instance, if $i^{th}$ and $j^{th}$ features are jointly permuted, the values are sampled from $p(x_{i,j}|x_{-i,j})$ by fixing all features other than $i,j$ and choose $x_{i,j}$ value from samples sharing similar discretized values in all features other than $i,j$. 

However, we found this to be a very loose estimation. To increase the accuracy of the desired joint feature sampling we use Gibbs sampling \cite{ait2009handbook}. To estimate a sample of a joint distribution, Gibbs sampling iteratively samples from the conditional distributions. It could be shown that this procedure converges to the joint distribution. In practice, this convergence happens in few iterations. Here, we use 5 iteration for each estimation. 

This sampling strategy can directly be used in \textit{mutation}. However, in crossover we swap the values of randomly selected features in the parents. To make sure that this is close enough to the real distribution of the data, we follow the same proposed sampling strategy but instead of randomly choosing from the possible choices, we choose the value closest to the one we want to swap. 

\section{Experiments on German Credit Data}

To test the proposed algorithm, \textit{German Credit} dataset is used \cite{germancredit}. The goal of the model is to predict the probability of \textit{defaulting} of a loan given the account and customer information. The data contains 20 features with 13 categorical ones, such as sex, purpose of the loan, housing type, if the customer owns a phone, if the customer is a foreign worker. A scikit-learn RandomForest model with 10 decision trees are used. The categorical features are one-hot encoded, creating 61 features in total. There are 1000 samples in total, which are splitted into 60\% train and 40\% test samples.

\begin{figure}[h]
	\centering
	\includegraphics[width=7cm]{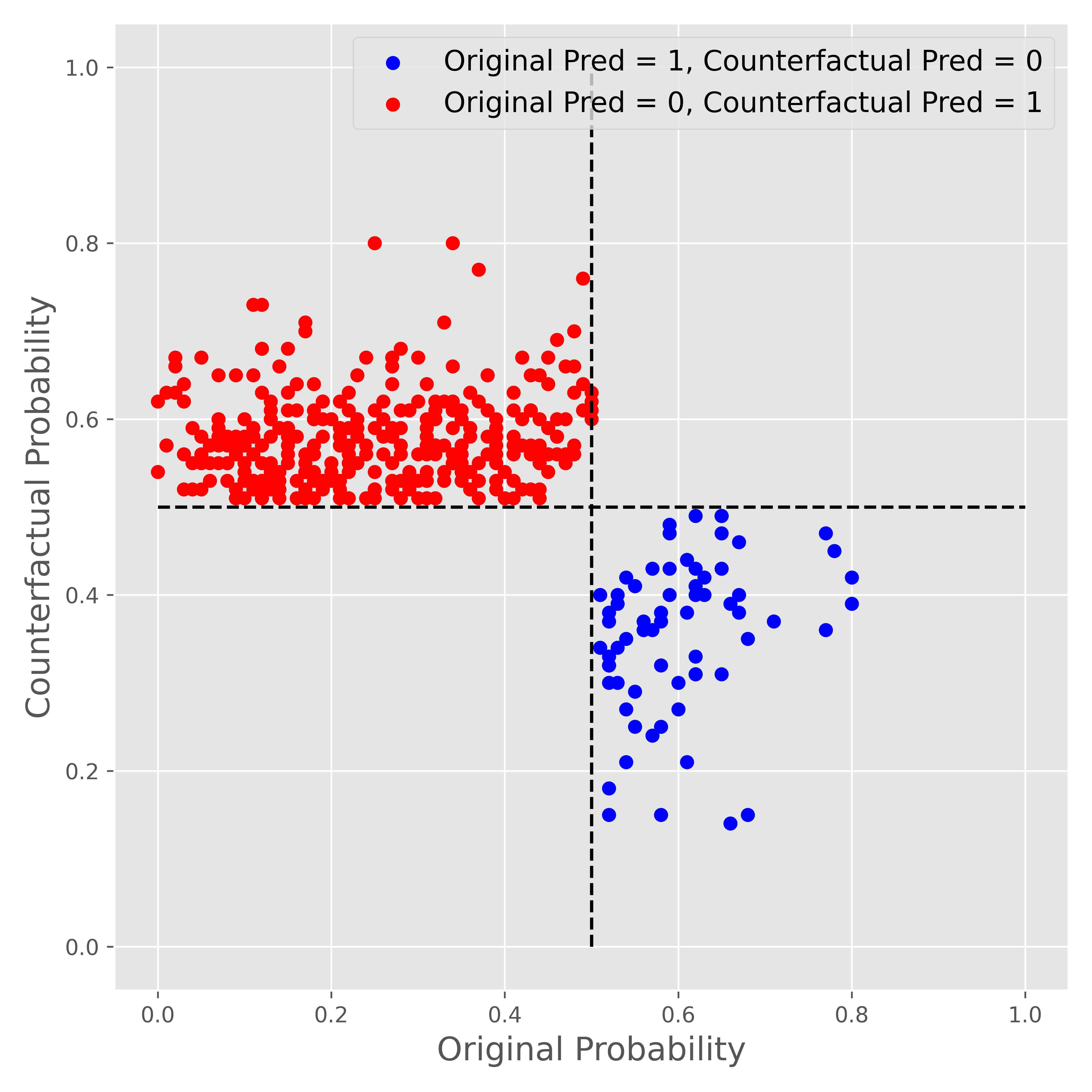}
	\caption{Shows how the outcomes are changed in the counterfactual examples. The x-axis shows the prediction outcome of the model for original examples. The y-axis shows the predictions for counterfactual examples.}
	\label{fig:scatteroutcomes}
\end{figure}

After training the model, PermuteAttack is applied on the test data. We use $\rho_0=0.6$ and $\rho_1=0.2$. Number of generation is 100, number of parents in mutations is 35, number of mating parents is 15, and sampling temperature is 0.5. In addition, if the probability of the outcome does not change in an iteration, $\rho_0$ and $\rho_1$ are relaxed by multiplication by 0.96. Using this hyper-parameters, the accuracy of the model drops to 0 after attack, meaning 100\% of the samples are successfully modified. The changes in the prediction of the model are depicted in Figure \ref{fig:scatteroutcomes}. All examples fall into top-left and bottom-right regions, which shows 100\% success in creation of counterfactual examples with different prediction classes. 

\begin{figure}[h]
	\centering
	\includegraphics[width=9cm]{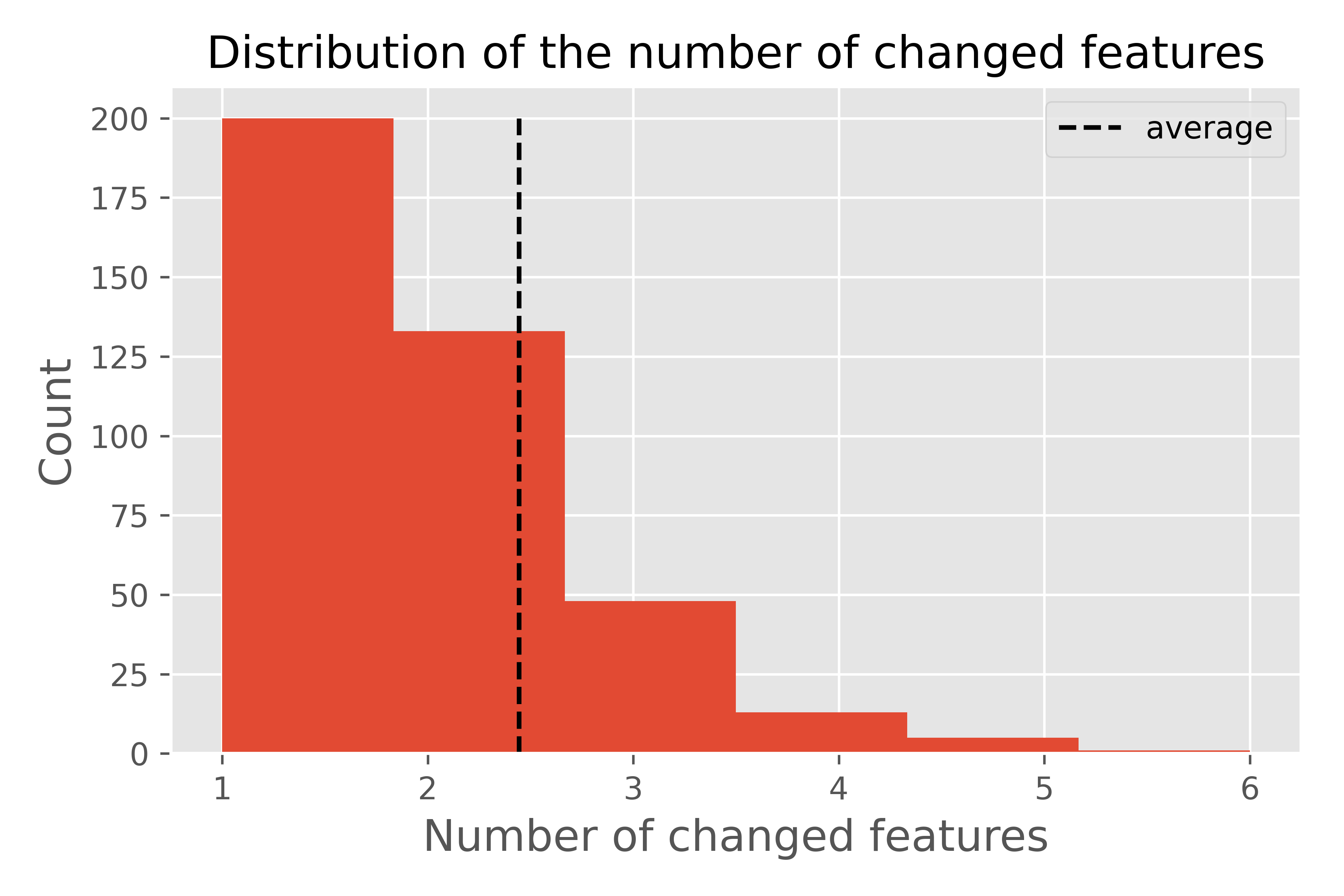}
	\caption{Histogram of the number of features that are changed in test samples of the German Credit data by PermuteAttack.}
	\label{fig:changefreq}
\end{figure}

\begin{figure}[h]
	\centering
	\includegraphics[width=11cm]{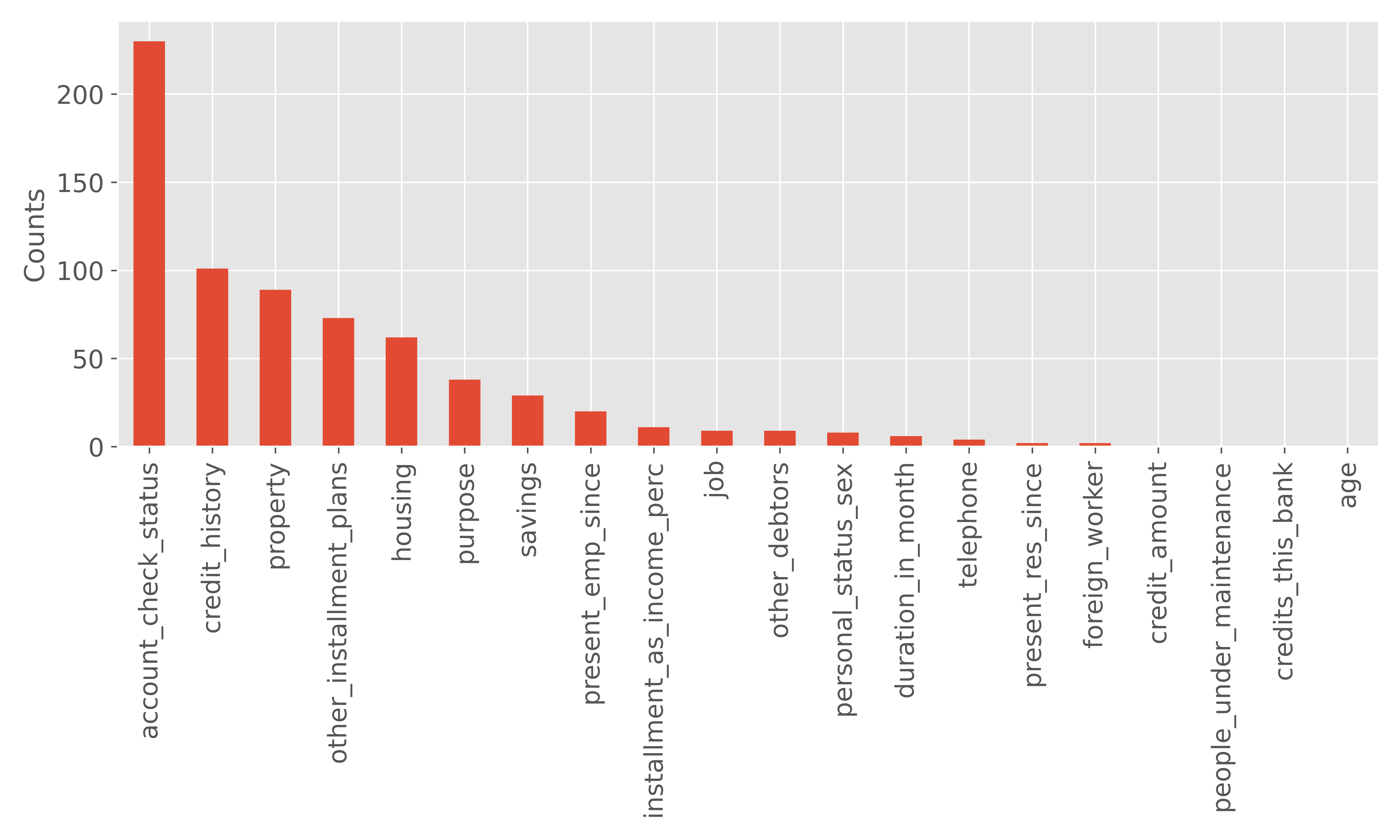}
	\caption{Number of samples that each feature has been changed in.}
	\label{fig:featchange}
\end{figure}

On average 2.35 features are changed to switch the German credit model predictions. Figure \ref{fig:changefreq} shows the distribution of the number of changed features for 400 test samples of the German credit data.  Figure \ref{fig:featchange} shows the number of samples that each feature has been changed in. 

Table \ref{tab:counterexample_neg} and \ref{tab:counterexample_pos} show a defaulting and a not-defaulting customer and their counterfactual examples generated with PermuteAttack. equation \ref{eqn:creditscore} is used to calculate the reported credit scores. 

The example shown in table \ref{tab:counterexample_neg} corresponds to a not-defaulting customer. Three counterfactual examples are created as the most similar defaulting cases. In the first example (CF1) ``purpose'' is changed from ``domestic appliances'' to ``new car'' and ``property'' is changed from ``saving / life insurance'' to ``unknown''. In CF2, property type is changed similar to CF1 and ``housing'' is changed from ``own'' to ``for free''. In CF3, housing is changed similar to CF2, job is changed from ``un-skilled resident'' to ``skilled employee official'', and ``present employee'' changed from ``1 to 4 years'' to ``less than 1 year''. 

The example shown in table \ref{tab:counterexample_pos} corresponds to a defaulting customer, for which, three not-defaulting counterfactual examples are reported. In CF1 ``saving'' is changed from ``$<$ 100 DM'' to ``$>$100DM''. In CF2, ``account\_check\_status'' is changed from ``$<$0 DM'' to ``$>$200 DM / salary assignments for at least 1 year''. In CF3 ``other\_debtors'' is changed from ``none'' to ``guarantor''. 

Reporting multiple alternatives as the counterfactual examples helps us to use the most relevant scenario for the customers when explaining the decision of the model and providing feedback to improve their credit score \cite{Wachter2017CounterfactualEW}. For instance in the reported example, to improve the credit score and to become eligible for the loan, the customer can increase the amount of saving, or use the checking account in the bank for salary deposits, or the customer can have a guarantor. 

\begin{table}[h!]
	\centering
\begin{tabular}{l || r | rrr}
	\toprule
	Features &        Original &        CF1 &       CF2 &       CF3 \\
	\midrule
	duration\_in\_month          &    12 &    12 &    12 &    12 \\
	credit\_amount              &  2214 &  2214 &  2214 &  2214 \\
	installment\_as\_income\_perc &     4 &     4 &     4 &     4 \\
	present\_res\_since          &     3 &     3 &     3 &     3 \\
	age                        &    24 &    24 &    24 &    24 \\
	credits\_this\_bank          &     1 &     1 &     1 &     1 \\
	people\_under\_maintenance   &     1 &     1 &     1 &     1 \\
	account\_check\_status       &     1 &     1 &     1 &     1 \\
	credit\_history             &     3 &     3 &     3 &     3 \\
	purpose                    &     4 &     \textbf{2} &     4 &     4 \\
	savings                    &     1 &     1 &     1 &     1 \\
	present\_emp\_since          &     2 &     2 &     2 &     \textbf{1} \\
	personal\_status\_sex        &     3 &     3 &     3 &     3 \\
	other\_debtors              &     2 &     2 &     2 &     2 \\
	property                   &     0 &     \textbf{3} &     \textbf{3} &     0 \\
	other\_installment\_plans    &     1 &     1 &     1 &     1 \\
	housing                    &     1 &     1 &     \textbf{0} &     \textbf{0} \\
	job                        &     3 &     3 &     3 &     \textbf{1} \\
	telephone                  &     0 &     0 &     0 &     0 \\
	foreign\_worker             &     1 &     1 &     1 &     1 \\
	\midrule
	Pred. Proba                &     0.13 &     0.59 &     0.6 &     0.6 \\
	Credit Score               &     641  &     592  &     591 &     591 \\
 	\bottomrule
\end{tabular}
\caption{Counterfactual examples for a not defaulting customer.}
\label{tab:counterexample_neg}
\end{table}

\begin{table}[h!]
	\centering
\begin{tabular}{l || r | rrr}
	\toprule
	Features &        Original &        CF1 &        CF2 &        CF3 \\
	\midrule
	duration\_in\_month          &    12 &    12 &    12 &    12 \\
	credit\_amount              &  1858 &  1858 &  1858 &  1858 \\
	installment\_as\_income\_perc &     4 &     4 &     4 &     4 \\
	present\_res\_since          &     1 &     1 &     1 &     1 \\
	age                        &    22 &    22 &    22 &    22 \\
	credits\_this\_bank          &     1 &     1 &     1 &     1 \\
	people\_under\_maintenance   &     1 &     1 &     1 &     1 \\
	account\_check\_status       &     1 &     1 &     \textbf{2} &     1 \\
	credit\_history             &     3 &     3 &     3 &     3 \\
	purpose                    &     7 &     7 &     7 &     7 \\
	savings                    &     1 &     \textbf{0} &     1 &     1 \\
	present\_emp\_since          &     1 &     1 &     1 &     1 \\
	personal\_status\_sex        &     0 &     0 &     0 &     0 \\
	other\_debtors              &     2 &     2 &     2 &     \textbf{1} \\
	property                   &     1 &     1 &     1 &     1 \\
	other\_installment\_plans    &     1 &     1 &     1 &     1 \\
	housing                    &     2 &     2 &     2 &     2 \\
	job                        &     1 &     1 &     1 &     1 \\
	telephone                  &     0 &     0 &     0 &     0 \\
	foreign\_worker             &     1 &     1 &     1 &     1 \\
	\bottomrule
	Pred. Proba                &     0.63 &     0.46 &     0.42 &     0.47 \\
	Credit Score               &     588  &     603  &     607 &     602 \\
	\bottomrule
\end{tabular}
\caption{Counterfactual examples for a defaulting customer.}
\label{tab:counterexample_pos}
\end{table}

\begin{figure}[h]
	\centering
	\includegraphics[width=11cm]{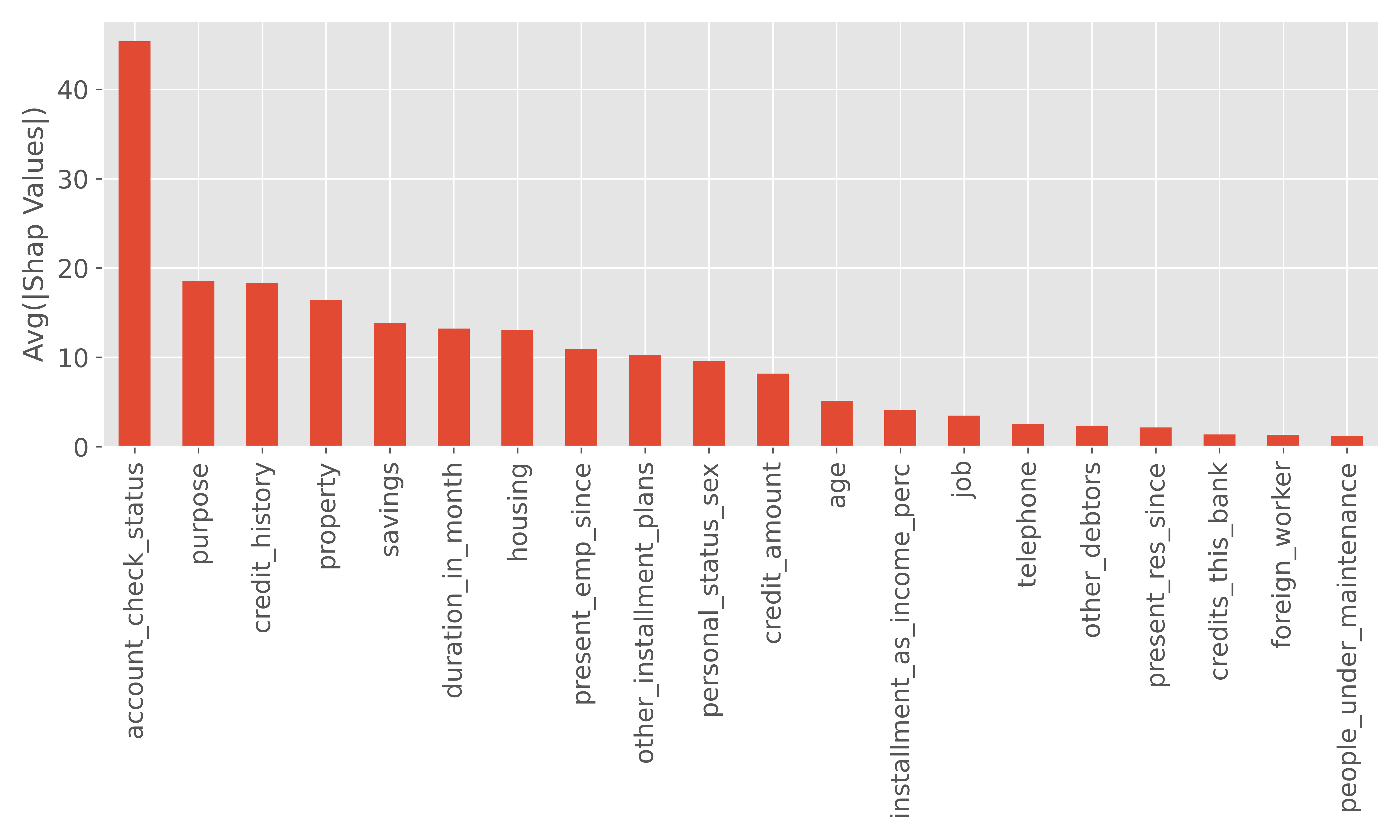}
	\caption{SHAP value feature importance \cite{lundberg2020local}.}
	\label{fig:germancredit_featimportance}
\end{figure}

To check if there is any relation between the adversarially perturbed features and the importance of the features in the model, SHAP feature importance is used. Figure \ref{fig:germancredit_featimportance} illustrates the feature importance of the trained model measured by SHAP values \cite{lundberg2020local}.

As can be seen from Figure \ref{fig:featchange} and Figure \ref{fig:germancredit_featimportance}, the most important features and the most frequently perturbed features are very well aligned. However, the exact order is changed in some features. For instance, ``other\_installment\_plans'' is the 10th important feature and 4th frequently perturbed feature. The same behaviour could be observed for ``housing''. On the other hand some features like ``duration\_in\_month'' are more important from SHAP perspective than adversarial perturbation. To understand the potential reason for this behaviour we can look at the features that are perturbed together. 

\begin{figure}[h]
	\centering
	\includegraphics[width=9cm]{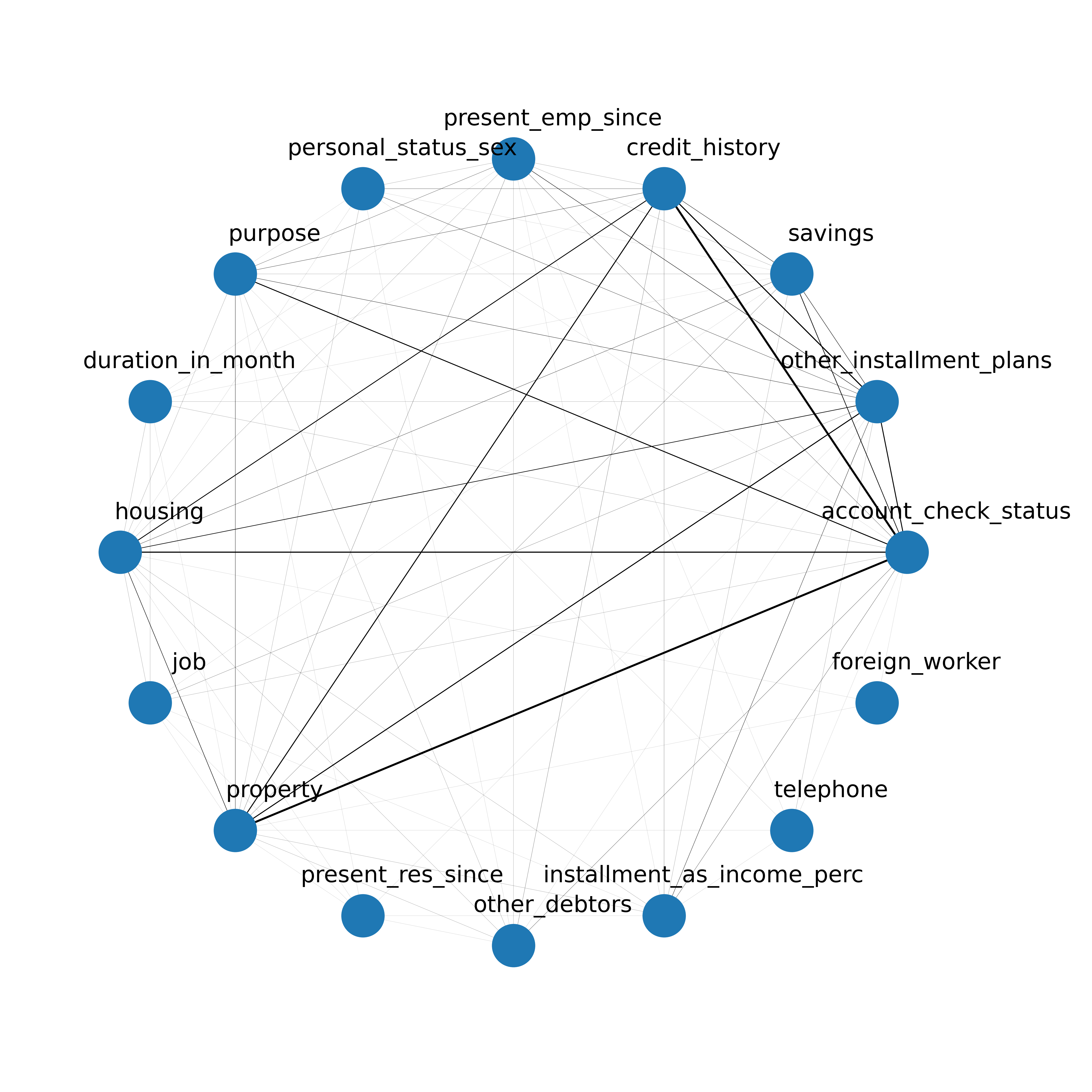}
	\caption{Graph of the co-occurrence of the changed features in German credit dataset.}
	\label{fig:germancredit1}
\end{figure}

Figure \ref{fig:germancredit1} shows how the features are adversarially perturbed together.  The vertices of the graph show the features that are changed together at least in one counterfactual example. The edges connect the vertices that are changed in the same counterfactual example. The edge width (connection weight) is proportional to the number of counterfactual examples that the edge is present in. For instance ``property'' and ``account\_check\_status'' are changed together in 42 examples, while ``job`` and ``account\_check\_status'' are change together in only 2 examples. Therefore, as can be seen in the graph the weight of the edge connecting ``property'' to ``account\_check\_status'' is much higher than the edge between ``job'' and ``account\_check\_status''. 

Looking at the degree of the features/vertices, it can be seen that ``other\_installment\_plans'' has a high degree with some strong connections to ``credit\_history'', ``property'', and ``account\_check\_status'', all of which are among the most important features of the model. Therefore, it could be hypothesized that there is a high interaction among these features in the model that is being utilized by PermutateAttack to flip the classification outcome. 

\begin{figure}[h]
	\centering
	\includegraphics[width=9cm]{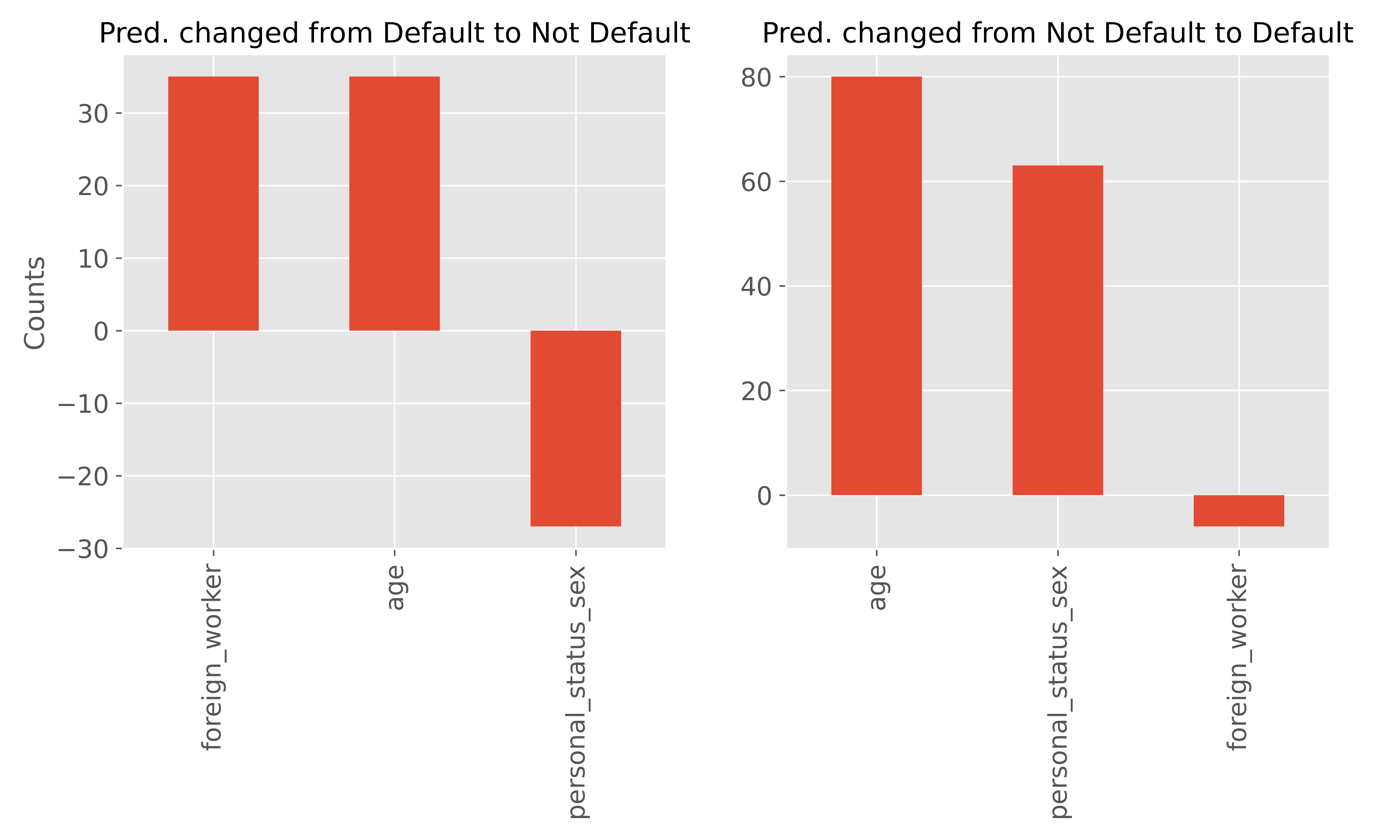}
	\caption{Aggregate of feature changes in 105 successful attacks by changing only the sensitive features, ``age'', ``personal\_status\_sex'', and ``foreign\_worker''.}
	\label{fig:sensitive_germancredit}
\end{figure}

Looking at the perturbed feature frequency, it can be observed that some information such as ``age'', ``personal\_status\_sex'', and ``foreign\_worker'' are not being utilized by the algorithm to change the outcome of the model. Does this mean that for people with same banking history, changing the gender and citizenship status does not have a significant effect on the outcome of the model? This can motivate more investigations about the model fairness. To do so, we use PermuteAttack to only perturb the sensitive attributes, namely: ``age'', ``personal\_status\_sex'', and ``foreign\_worker''. Using these three features 26.25\% of the attacks are successful (105 out of 400 test samples). Figure \ref{fig:sensitive_germancredit} shows the aggregate feature perturbations when the predictions are changed from default to not-defaulting and vice-versa. Based on this result, changing ``personal\_status\_sex'' could cause a change in prediction from default to not-default; and changing ``foreign\_worker'' from \textit{yes} to \textit{no} could change the prediction from not-default to default. However, there is an interaction among these three features, causing non-monotonicity in the model, that makes the exact analysis more complicated. For example in both cases the overall increase in age is more effective than its reduction, with a more pronounced effect on not-default to default changes. The overall hypothesis would be that there are possible discrimination against some of the sensitive attributes.

In \cite{ballet2019imperceptible} it is suggested that an adversarial attack on tabular data is \textit{imperceptible} if attackers minimize the manipulations on the important features, which are easily detectable by an expert. Therefore, the attack to be imperceptible should rely on less important features which are harder to detect. 

\begin{figure}[h]
	\centering
	\includegraphics[width=9cm]{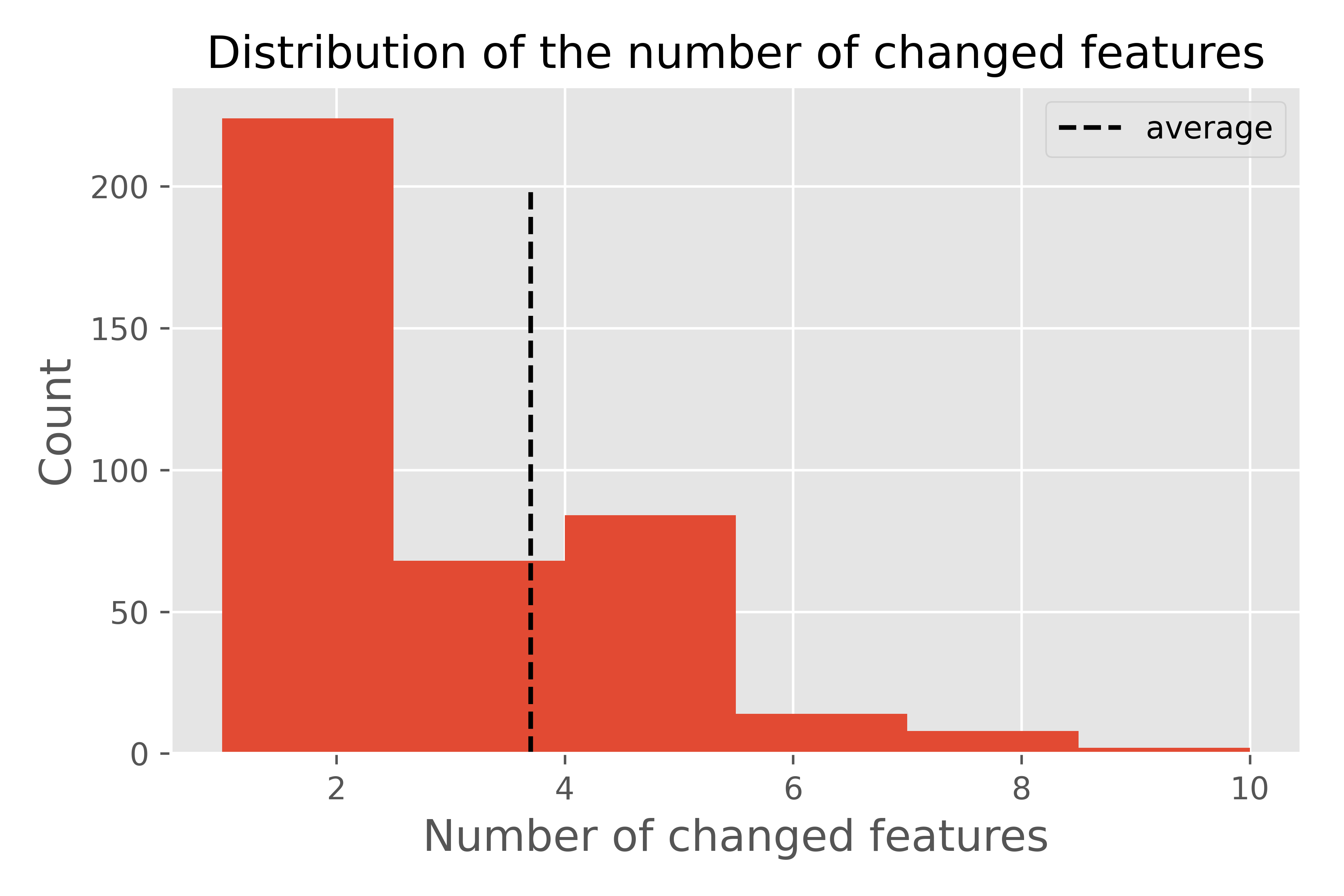}
	\caption{Histogram of the number of features that are changed in test samples of the German Credit data by PermuteAttack when most important features (based on SHAP values) are removed.}
	\label{fig:chenge_freq_impremoved}
\end{figure}

\begin{figure}[h]
	\centering
	\includegraphics[width=11cm]{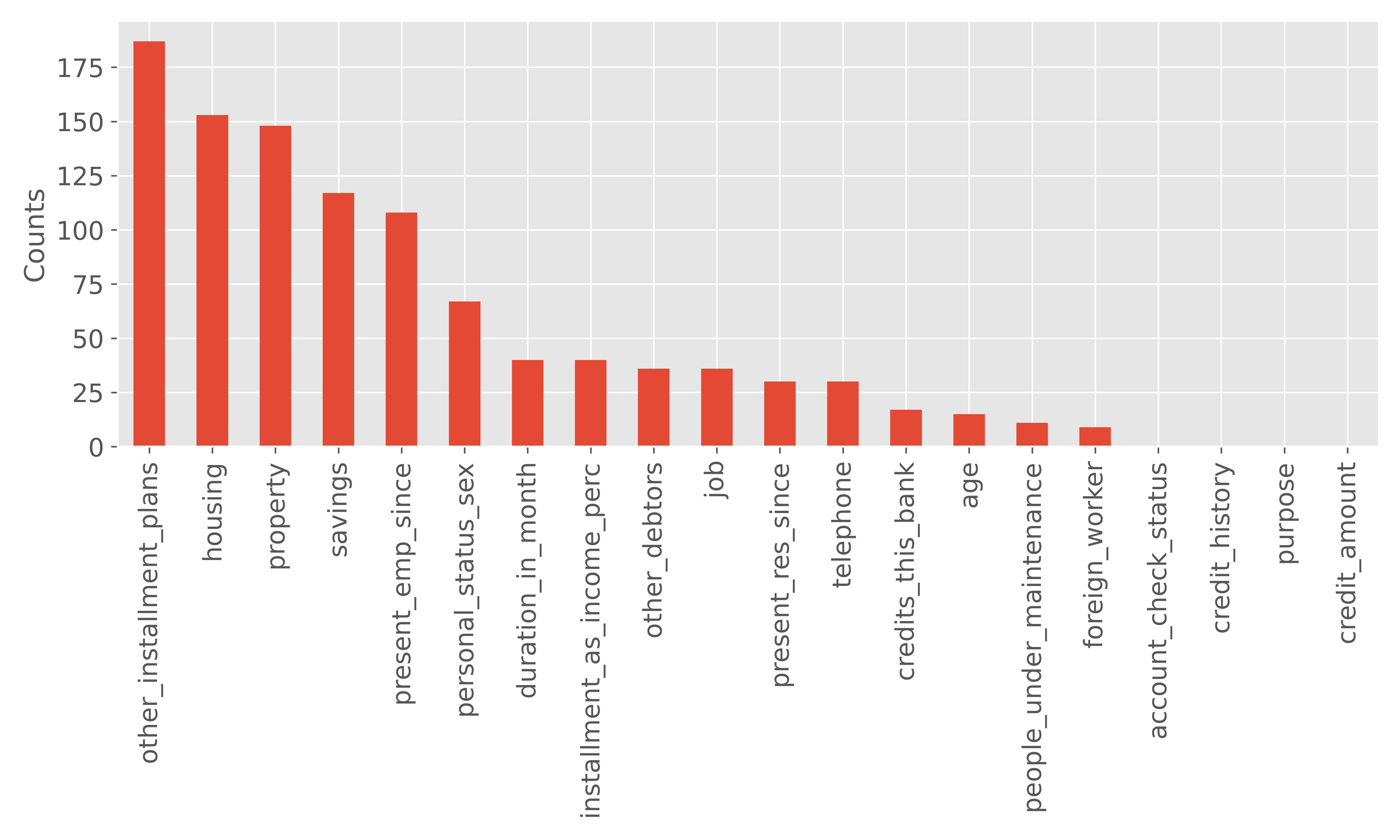}
	\caption{Number of samples that each feature has been changed in, after removing the most important features, based on SHAP values.}
	\label{fig:feat_change_hist_impremoved}
\end{figure}

\begin{figure}[h]
	\centering
	\includegraphics[width=9cm]{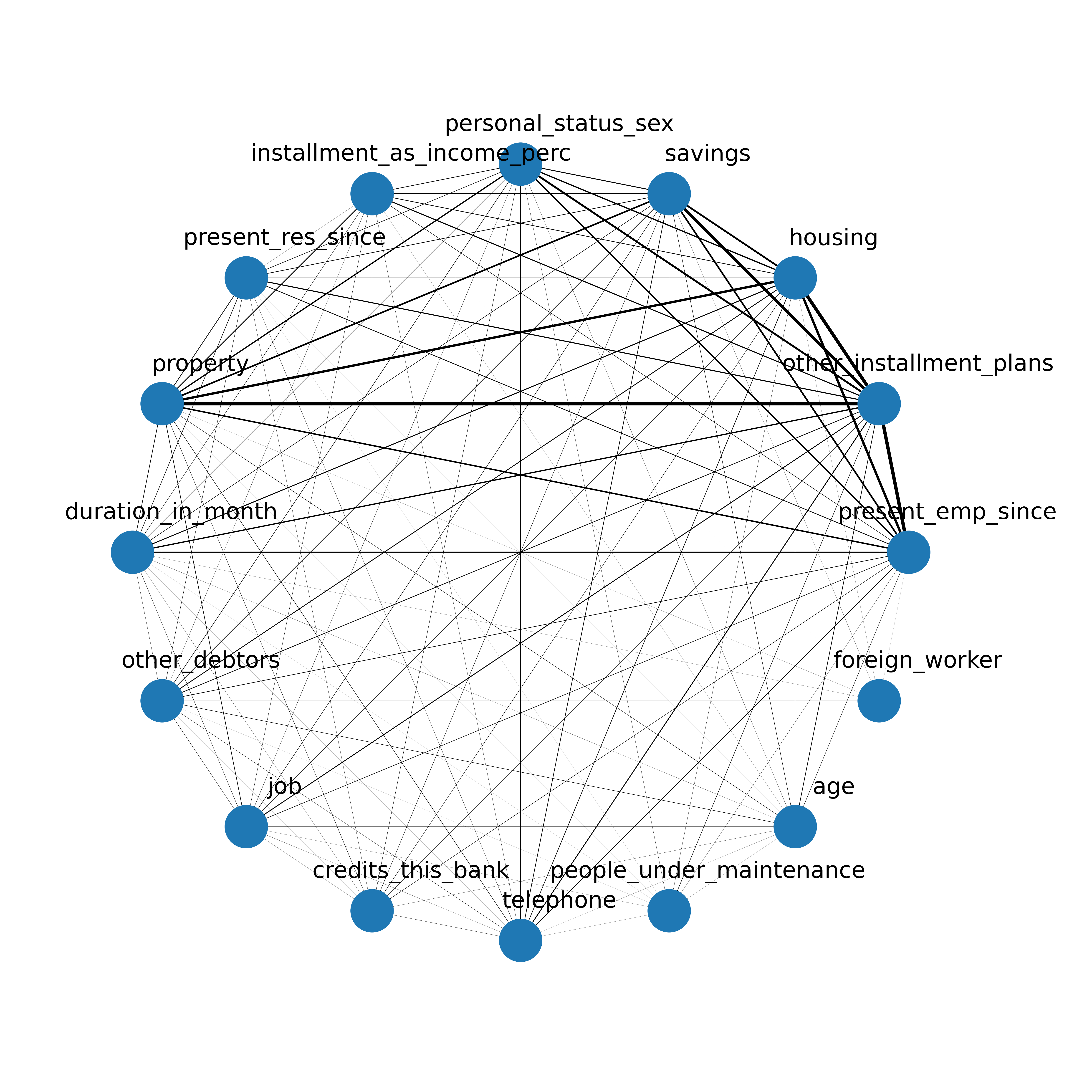}
	\caption{Graph of the co-occurrence of the changed features in German credit dataset, after removing the most important features, based on SHAP values.}
	\label{fig:featchange_graph_impremoved}
\end{figure}

Although, we use PermuteAttack as an algorithm for counterfactual example generation, since it is based on the idea of adversarial attack, we decided to explore the effect of removing the most important features from attack. The feature importance is measured by SHAP, Figuer \ref{fig:germancredit_featimportance}; and the first three most important features,  ``account\_check\_status'', ``purpose'', and ``credit\_history'' are excluded from PermuteAttack. As depicted in Figure \ref*{fig:chenge_freq_impremoved}
after excluding the most important features, on average the number of features that should be changed to successfully change the prediction increases to 3.7. A summary of the perturbed features is shown in Figure \ref{fig:feat_change_hist_impremoved}. Figure \ref{fig:featchange_graph_impremoved} shows the co-occurrence of the features.

To check the effect of Gibbs sampling in making more realistic counterfactual examples, we train a model to discriminate between the real and adversarially generated samples. To do so we run PermuteAttack on the test data and generate one set of counterfactual examples. The real data is labeled one and the counterfactual examples are labeled zero. Using this data a model is trained. Then we change the random seed and generate another set of new counterfactual examples to test the trained model. The more similar the generated examples are to the real data, the accuracy of the test will be lower. 

We trained two models: one using the naive method (without Gibbs sampling) and one with Gibbs sampling. Each model is tested on the counter examples generated using the same technique used to generate the training data. 

The model trained on counter-examples without Gibbs sampling fails in 12\% of the test counter-examples. This means that only 12\% of the generated counter-examples are similar enough to the real data to not be detected by discriminator model as adversarial examples. However, when the model is trained with the examples generated with Gibbs sampling it fails in 49\% of the test samples. Meaning that 49\% of the generated counterexamples are similar enough to real data to not be detected with the discriminator model as adversarially generated samples. 

\section{Conclusions}
In this paper PermuteAttack was proposed as a model criticism and explanation algorithm  based on \textit{adversarially generated counterfactual examples} for validation and explanation of Machine Learning (ML) models employed for retail credit scoring in finance trained on tabular data. Using the counterfactual examples increases the model behaviour understanding and can offer helpful feedback to the customers to improve the credit scores. PermuteAttack generates sensible and realistic counterfactual examples using permutation as the adversarial perturbation which keeps the range and the distribution of each individual feature the same as the original training data. In addition, using Gibbs sampling the joint distribution of the synthetically generated examples follow the original data distribution. 

\section{Future Directions}
In this paper data discretization and Gibbs sampling are used to estimate conditional probability of the features. A possible future direction would be replacing this process with more accurate techniques such as conditional Gaussian Copula models \cite{malevergne2003testing} and conditional generative machine learning models \cite{yoon2018gain}. 

\begin{figure}[h]
	\centering
	\fcolorbox{black}{white}{\includegraphics[width=0.8\linewidth]{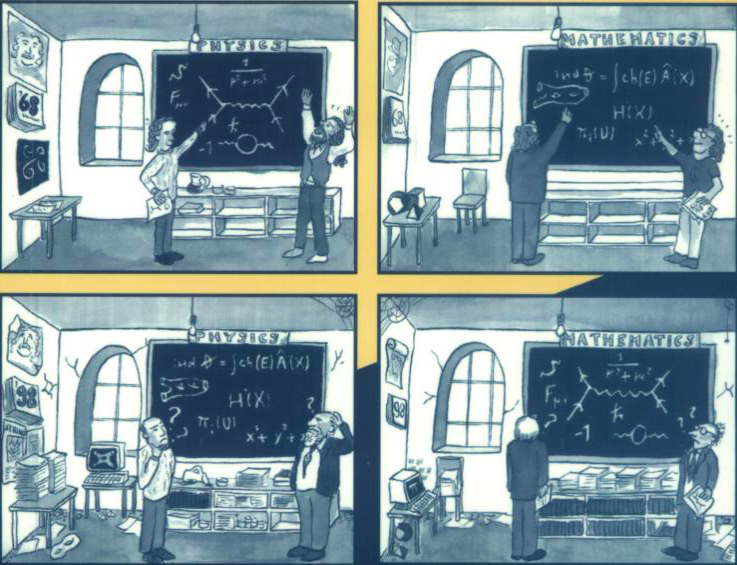}}
	\caption{The authors are fans of the cartoon on the cover page of \cite{Delign1999QuantumFA},  describing one of the exciting interactions between theoretical physics and mathematics in the past fifty years.(Top row- in the 60's), left: physicists studying Feynman diagrams- considered as ill-defined objects by mathematicians, right: mathematicians inventing Index Theory. (bottom row- in the 90's), folks pondering the same notions from the lens of the other camp. The cartoon fits to the narrative of Brieman's ``Two Cultures'' \cite{Breiman2001StatisticalMT}  on fruits of the clashes between statistics and machine learning camps as scientific disciplines.}
	\label{qft_cover}
\end{figure}

\bibliography{ref}
\bibliographystyle{unsrt}

\end{document}